\documentclass[11pt,a4paper]{article}
\usepackage[hyperref]{emnlp-ijcnlp-2019}
\usepackage{times}
\usepackage{latexsym}
\usepackage{graphicx}
\usepackage{lipsum}
\usepackage{xspace}
\usepackage{multicol}
\usepackage{graphicx}
\usepackage{mathrsfs}
\usepackage{multirow}
\usepackage{caption}
\usepackage{amssymb}
\usepackage{changepage}
\usepackage{amsmath}
\usepackage{tabularx}
\usepackage{comment}

\usepackage{url}

\newcommand\numberthis{\addtocounter{equation}{1}\tag{\theequation}}

\aclfinalcopy 

\newcommand{\onetoone}{\textbf{One2One}\xspace}
\newcommand{\onetoseq}{\textbf{One2Seq}\xspace}

\newcommand{\nosort}{\textbf{\texttt{\textsc{No-Sort}}}\xspace}
\newcommand{\random}{\textbf{\texttt{\textsc{Random}}}\xspace}
\newcommand{\alphab}{\textbf{\texttt{\textsc{Alpha}}}\xspace}
\newcommand{\length}{\textbf{\texttt{\textsc{Length}}}\xspace}
\newcommand{\appearpre}{\textbf{\texttt{\textsc{Appear-Pre}}}\xspace}
\newcommand{\appearap}{\textbf{\texttt{\textsc{Appear-Ap}}}\xspace}

\newcommand{\kpk}{\texttt{\textsc{KP20k}}\xspace}

\newcommand{\duc}{\texttt{\textsc{DUC}}\xspace}
\newcommand{\inspec}{\texttt{\textsc{Inspec}}\xspace}
\newcommand{\krapivin}{\texttt{\textsc{Krapivin}}\xspace}
\newcommand{\nus}{\texttt{\textsc{NUS}}\xspace}
\newcommand{\semeval}{\texttt{\textsc{SemEval}}\xspace}

\newcommand{\smx}{{\mathrm{softmax}}}

\definecolor{color1}{HTML}{da6752}
\definecolor{color2}{HTML}{5573a6}
\definecolor{color3}{HTML}{6f9f6a}
\definecolor{color4}{HTML}{f3905c}

\title{Does Order Matter? An Empirical Study on Generating Multiple Keyphrases as a Sequence}

\author{Rui Meng$^\clubsuit$ \:\:\:\: Xingdi Yuan$^\spadesuit$ \:\:\:\: Tong Wang$^{\spadesuit}$ \\ 
\textbf{Peter Brusilovsky$^{\clubsuit}$ \:\:\:\: Adam Trischler$^{\spadesuit}$ \:\:\:\: Daqing He$^{\clubsuit}$}\\
$^\clubsuit$School of Computing and Information, University of Pittsburgh \\
$^\spadesuit$Microsoft Research, Montr\'{e}al \\
rui.meng@pitt.edu \:\:\:\: eric.yuan@microsoft.com
}

\date{}

\begin{document}
\maketitle

\noindent\textcolor{red}{Update: this work has been expanded to a more general study on the keyphrase generation task with more comprehensive experiments and analyses on 1) generalization, 2) target phrase ordering, and 3) pre-training with noisy annotation. Please refer to the latest version at \url{https://arxiv.org/abs/2009.10229}.}

\begin{abstract}
Recently, concatenating multiple keyphrases as a target sequence has been proposed as a new learning paradigm for keyphrase generation. 
Existing studies concatenate target keyphrases in different orders but no study has examined the effects of ordering on models' behavior. 
In this paper, we propose several orderings for concatenation and inspect the important factors for training a successful keyphrase generation model. 
By running comprehensive comparisons, we observe one preferable ordering and summarize a number of empirical findings and challenges, which can shed light on future research on this line of work.
\end{abstract}

\section{Introduction \& Related Works}
\label{section:intro} 
Keyphrases are multi-word units used for summarizing high-level meaning of a longer text and highlighting certain important topics or information. 
As an important capacity of language understanding and knowledge extraction systems, keyphrase extraction has been discussed for over a decade \citep{witten1999kea, liu2011gap, wang2016ptr, yang2017semisupervisedqa, luan2017tagging,subramanian2017kp, sun19divgraphpointer}.
However, extractive models lack the ability to predict keyphrases that are absent from the source text (i.e., keyphrases that abstract and summarize key ideas of the source text).
\citet{meng2017deep} first propose CopyRNN, a neural model that both generates words from a vocabulary and copies words from the source text. It has since served as the basis for a host of later works \citep{chen2018kp_correlation,ye2018kp_semi,yuan2018diversekp,cano19struggle}.


Given a piece of source text, our objective is to generate a set of multi-word phrases. This falls naturally under the paradigm of set generation (i.e., output phrases should be permutation invariant). 
There exists some literature exploring the effects of using different orderings in the task of language modeling \citep{vinyals2016_order_matters, ford18orderlm}. 
Recently, \citet{yang18seq2set} also propose a sequence-to-set model trained with Reinforcement Learning, which captures the correlation between labels and reduces the dependence on label order for the multi-label text classification.

To our best knowledge, however, there is no existing work that has successfully applied set generation to real-world, large scale language generation tasks such as keyphrase generation. Compared to multi-class classification tasks, the complexity in keyphrase generation is combinatorially larger --- each phrase is a multi-word sequence with a usually very large vocabulary.

To this end, most existing keyphrase generation methods aim to generate a single keyphrase for each source text during training. During decoding, beam search is often used to produce a large number of candidate phrases.
However, this decoding strategy generates all target phrases independently, resulting in many similar or identical phrases being generated.
Furthermore, as an intrinsic limitation of beam search, a single beam may dominate the search process and further diminish the diversity of the final output.

Many studies have noted this problem \citep{chen2018kp_correlation, ye2018kp_semi}. 
Consequently, instead of independent generation, another line of work proposes to generate multiple phrases in one output sequence \citep{yuan2018diversekp}, where models are trained to generate the concatenation of target phrases.
While overcoming the problem of independent generation, this latter approach introduces a new question of ordering among the now inter-dependent phrases, as pointed out by~\cite{vinyals2016_order_matters}, order matters for sequence modeling. However, previous studies have largely overlooked this problem. 


In this study, we aim to fill this research gap by systematically examine the influence of concatenation ordering, as well as other factors like beam width and model complexity, to sequential generation models for keyphrase generation. By conducting comprehensive empirical experiments, we find our model delivers superior performance, indicating that learning to generate multiple phrases as a sequence is an effective paradigm for this task. More importantly, models trained with certain orderings consistently outperform others.

\section{Generating Multiple Keyphrases as a Sequence}
\label{section:method}


\subsection{Model Architecture}
\label{section:structure}

In this paper we use \onetoone to denote the training and decoding strategy where each source text corresponds to a single target keyphrase, and a common practice is to use beam search to over-generate multiple keyphrases. This is in contrast of \onetoseq, where each source text corresponds to a sequence of keyphrases that are concatenated with a delimiter token \texttt{\textsc{<SEP>}}\xspace. By simple greedy search, a model with \onetoseq setting is capable of generating a sequence of multiple phrases. But the over-generation with beam search is often necessary to boost the recall. Please refer to~\cite{yuan2018diversekp, ye2018kp_semi} for details.

We adopt a sequence-to-sequence based framework with pointer-generator and coverage mechanism proposed by \citet{see17gettothepoint}. Our focus in this work is to study what factors are most critical for models trained with \onetoseq setting, such as keyphrase ordering and beam width, rather than the model structure itself, thus we describe details of the model structure in Appendix~\ref{appd:structure}.

\subsection{Ordering for Concatenating Phrases}
In this subsection, we define six ordering strategies for concatenating target phrases as follows. We are interested in seeing if different orderings affect the performance on keyphrase generation and which orderings may be optimal for training models in the \onetoseq~setting.
\begin{itemize}
    \setlength\itemsep{-0.2em}
    \item \random: Randomly shuffle the target phrases. As the goal of keyphrase generation is to output an order-invariant structure (a set of phrases), we expect models trained with randomly shuffled targets would capture such nature better than other variants with more fixed ordering.
    \item \nosort: Keep phrases in original order. Also used by~\cite{ye2018kp_semi}.
    \item \length: Sort phrases by their lengths from short to long. Phrases of the same length are sorted in original order.
    \item \alphab: Sort phrases in alphabetical order (by their first word). 
    \item \appearpre: Sort present phrases by their first occurrences in the source text, and prepend absent phrases at the beginning. Absent phrases are randomly shuffled.
    \item \appearap: Same to \appearpre but append absent phrases at the end. Also used by~\cite{yuan2018diversekp}.
    
\end{itemize}


\subsection{Efficient Decoding Strategy}
\label{section:decoding}
Previous study~\cite{ye2018kp_semi} adopts beam search and a phrase-ranking technique to collect a excessive number of unique phrases. However in the setting of \onetoseq, this decoding strategy can cause very high computational cost, as a result of longer decoding sequence and much deeper beam search process. 

In order to make \onetoseq decoding more computationally affordable, we propose to use a early-stop technique during beam search. Instead of expanding all the search branches until reaching a given maximum depth, we terminate the beam search once the best sequence is found. This is a common heuristic for speed-up in single-sequence generation tasks such as translation and summarization. We observe that it is also effective for \onetoseq decoding, leading to up to 10 times faster decoding. By the time that top sequence is completed, there is usually enough number of good phrases, meanwhile the quality of later generated sequences degenerates drastically and most of them are duplicates of existing phrases. Therefore this early-stop technique achieves a significant efficiency gain without sacrificing the quality of output phrases.  

\begin{table*}[!htbp]
  \centering
  \fontsize{8}{10}\selectfont
  \renewcommand{\arraystretch}{1.2}
  \begin{tabular}{c|cc|cc|cc|cc|cc}
    \hline
    \hline
    \multirow{2}{*}{\textbf{Method}} & \multicolumn{2}{c|}{\textbf{Inspec}~\color{black}\tiny (Avg=7.8)
        }
    & \multicolumn{2}{c|}{\textbf{Krapivin}~\color{black}\tiny (Avg=3.4)
        }
    & \multicolumn{2}{c|}{\textbf{NUS}~\color{black}\tiny (Avg=6.1)
        }
    & \multicolumn{2}{c|}{\textbf{SemEval}~\color{black}\tiny (Avg=6.7)
        }
    & \multicolumn{2}{c}{\textbf{Average}
        }
    \\
    \cline{2-11}
     & \textbf{F$_1$@5}& \textbf{F$_1$@10} & \textbf{F$_1$@5} & \textbf{F$_1$@10}
    & \textbf{F$_1$@5} & \textbf{F$_1$@10} & \textbf{F$_1$@5} & \textbf{F$_1$@10} & \textbf{F$_1$@5} & \textbf{F$_1$@10}
    \\  
    \hline
    \hline
     \onetoone 
     & 0.244 & 0.289 & 0.305 & \textbf{0.266} & \textbf{0.376} & \textbf{0.352} & 0.318 & \textbf{0.318} & 0.311 & \textbf{0.306}
    \\  \hline \hline
     \random 
    & 0.283 & 0.206 & 0.288 & 0.183 & 0.344 & 0.238 & 0.304 & 0.218 & 0.305 & 0.211
    \\ \hline
	 \length 
    & 0.298 & 0.224 & 0.321 & 0.206 & 0.364 & 0.259 & 0.311 & 0.222 & 0.324 & 0.228
    \\ \hline
	 \nosort 
	& 0.323 & 0.253 & 0.317 & 0.209 & \textbf{0.376} & 0.264 & 0.318 & 0.244 & 0.333 & 0.243
    \\ \hline
	 \alphab 
	& 0.319 & 0.283 & \textbf{0.329} & 0.238 & \textbf{0.376} & 0.289 & \textbf{0.343} & 0.266 & \textbf{0.342} & 0.269
    \\  \hline
	 \appearpre 
	& 0.320 & 0.307 & 0.322 & 0.245 & 0.369 & 0.302 & 0.327 & 0.291 & 0.334 & 0.286
    \\  \hline
	 \appearap 
	& \textbf{0.344} & \textbf{0.333} & 0.320 & 0.236 & 0.367 & 0.295 & 0.324 & 0.286 & 0.339 & 0.287
    \\  
    \hline\hline
  \end{tabular}
  \caption{Present keyphrase prediction performance ($F_1$-score) on four benchmark datasets. Dataset names are listed in the header followed by the average number of target phrases per document. The baseline \onetoone~model is on the first row, followed by different \onetoseq~variants.  Bold indicates best score in each column.}
  \label{tab:performance-comparison}
\end{table*}

\section{Experiment Settings}
\label{section:setting}
Following the experiment setting in~\cite{meng2017deep}, we train all the models with the \kpk training set, which contains 514,154 scientific papers each with a title, an abstract and a list of keyphrases provided by the author(s). We take four common datasets \inspec, \krapivin, \nus, and \semeval for testing. We report the testing performance of best checkpoints, which are determined by F1@5 score on validation
set, containing 500 data examples from the \kpk validation set. 

In both settings of \onetoseq and \onetoone, we use model described in \S\ref{section:structure}. In the \onetoseq setting, we experiment with beam widths of [10, 30, 50] in decoding phase; while in \onetoone, we use a beam width of 200 for fair comparison with \citep{meng2017deep}. Note that only unique phrases are used for evaluation.


\section{Results and Discussion}
\label{section:result}
We present and discuss the results of present phrases (i.e, appearing directly and verbatim in the source text) in \S\ref{subsec:ordering}-\ref{subsec:complexity}, and absent phrases in \S\ref{subsec:absent}.

\subsection{Effects of Ordering}
\label{subsec:ordering}

We report experimental results on four common benchmark datasets (totalling 1241 testing data points) in Table~\ref{tab:performance-comparison}. Our implementation of the \onetoone~model~\cite{meng2017deep} yields much better scores on \nus~and \semeval, and comparable performance on the remaining datasets. Meanwhile, \onetoseq~models produce top 5 phrases in better quality than the baseline and there is no noticeable difference on their $F_1$@5. However, the scores deteriorate if we include top 10 outputs. As shown in Table~\ref{tab:performance-comparison}, the average performance increases from \random~to \appearap: a trend that becomes particularly obvious for $F_1$@10. We will elaborate on this observation shortly and offer possible explanations. 

Table \ref{tab:pred-stats} presents the statistics on the output of various models. We can see a notable correlation between $F_1$-score and the number of unique predicted keyphrases (\#(UniqKP)). Specifically, both \random~and \length~predict less than 5 unique phrases on average, leading to lower $F_1$. Intuitively, we were expecting \random~to help capture the order-invariance among phrases. In practice, however, the random phrase order seems to have induced more difficulties than robustness in learning. Since the lengths of most phrases are in the range between 1 to 4, \length~ordering only presents a very weak clue for models to exploit, especially given that counting itself can be a rather challenging task. Similarly, despite the static nature of the phrase order in \nosort~is, there also exists a high degree of arbitrariness in the way different authors list keyphrases in different papers. Such intrinsic randomness again poses much difficulty in learning.

In contrast, \appearpre and \appearap have the largest number of unique predictions as well as the best scores on \textbf{$F_1$@10}. In particular, \appearap~out-numbers others in both number of beams and phrases by a large margin. We postulate that the order of the targets' occurrence in the source text provides the pointer (copying) attention with a fairly reliable pattern to follow. This is in line with the previous findings that using pointer attention can greatly facilitate the training of keyphrase generation~\cite{meng2017deep}. In addition, appending the absent phrases to the end of the concatenated sequence may cause less confusion to the model. Finally, the \alphab~ordering surprisingly yields satisfactory scores and training stability. We manually check the output sequences and notice that the model is actually able to retain alphabetical order among the predicted keyphrases, hinting that the model might be capable of learning simple morphological dependencies (sort words in alphabetical order) even without access to any character-level representation.

\begin{table}[!htbp]
  \centering
  \fontsize{9}{10}\selectfont
  \scriptsize
  \renewcommand{\arraystretch}{1.2}
  \begin{tabular}{c|cccc}
    \hline
    \hline
    &  \textbf{\#(Beam)}& \textbf{Len(Beam)} & \textbf{\#(UniqKP)} & \textbf{\#(KP)}
    \\  
    \hline
    \random 
    & 13.01 & 21.72 & 4.27 & 47.13
    \\
	\length 
    & 8.84 & 17.74 & 4.89 & 32.58
    \\
	\nosort 
    & 14.33 & 23.60 & 5.25 & 56.19
    \\
	\alphab 
    & 12.15 & 23.31 & 6.71 & 51.24
    \\ 
	\appearpre 
    & 13.09 & 19.05 & 7.60 & 45.55
    \\ 
	\appearap 
    & \textbf{23.73} & \textbf{28.86} & \textbf{7.82} & \textbf{128.32}
    \\ 
    \hline\hline
  \end{tabular}
  \caption{Statistics of predictions from various models. For each model variant, \#(Beam) and Len(Beam)  are the average number and average length of all predicted beams (BeamWidth=10). \#(UniqKP) means the average number of unique predicted keyphrases and \#(KP) is the average number of all predicted keyphrases.}
  \label{tab:pred-stats}
\end{table}

\subsection{Effects of Beam Width}
\label{subsec:beamwidth}
We report the model performance with different beam widths in Table \ref{tab:beam_width}. From the results we can see a noticeable trend that, regardless of the target order, all models have a significant performance boost with a bigger beam width, which is largely due to the larger number of unique predictions. However the performance gap among different orderings remains clear, indicating that training ordering has a consistent effect on \onetoseq~models. With beam width 50, \appearap~wins against \onetoone~on both $F_1$@5 and @10. But to a certain extent, more unique predictions may also introduce noise. Compared with the other four orderings, \appearap and \appearpre have much lower precision and $F_1$ scores for top 5 predictions with beam width 25 and 50.

\begin{table}[!htbp]
  \centering
  \fontsize{9}{10}\selectfont
  \scriptsize
  \renewcommand{\arraystretch}{1.2}
  \begin{tabular}{c|cc|cc|cc}
    \hline
    \hline
    \textbf{Beam Width} 
    & \multicolumn{2}{c|}{\textbf{10}}
    & \multicolumn{2}{c|}{\textbf{25}}
    & \multicolumn{2}{c}{\textbf{50}}
    \\
    \hline
    &  \textbf{F@5}& \textbf{F@10} & \textbf{F@5}& \textbf{F@10} & \textbf{F@5}& \textbf{F@10}
    \\  
    \hline
    \random 
    &  0.305 & 0.211 & 0.345 & 0.261 & 0.358 & 0.304
    \\
	\length 
    &  0.324 & 0.228 & 0.347 & 0.279 & 0.351 & 0.319
    \\
	\nosort 
    &  0.333 & 0.243 & \textbf{0.357} & 0.295 & \textbf{0.364} & 0.325
    \\
	\alphab 
    &  \textbf{0.342} & 0.269 & 0.349 & 0.320 & 0.354 & 0.341
    \\ 
	\appearpre 
    &  0.334 & 0.286 & 0.340 & 0.326 & 0.337 & 0.345
    \\ 
	\appearap 
    &  0.339 & \textbf{0.287} & 0.340 & \textbf{0.329} & 0.339 & \textbf{0.347}
    \\ 
    \hline\hline
  \end{tabular}
  \caption{Average $F_1$ scores on present keyphrase generation with different beam widths.}
  \label{tab:beam_width}
  \vspace{-1em}
\end{table}

\subsection{Effects of Model Complexity}
\label{subsec:complexity}
After concatenating multiple phrases, the target sequence of \onetoseq~models becomes much longer and it may require more parameters to model the dependency. Therefore we are interested in knowing whether one can achieve better performance by increasing model complexity. Besides the afore-mentioned base model (referred as \textbf{BaseRNN}), two larger models are used for comparison: (1) \textbf{BigRNN} used in~\cite{ye2018kp_semi}, same architecture as the base model except a larger embedding size (128) and hidden size (512); (2) \textbf{Transformer} used in~\cite{gehrmann2018_bottomup}, a four-layer transformer with 8 heads, 512 hidden units and the copy attention. As shown in Table \ref{tab:model_complexity}, neither of the two is able to outperform \textbf{BaseRNN}. Interestingly, the performance differences among six orderings are consistently observed on the bigger models.
\begin{table}[!htbp]
  \centering
  \fontsize{9}{10}\selectfont
  \scriptsize
  \renewcommand{\arraystretch}{1.2}
  \begin{tabular}{c|cc|cc|cc}
    \hline
    \hline
    \textbf{Model} 
    & \multicolumn{2}{c|}{\textbf{BaseRNN}}
    & \multicolumn{2}{c|}{\textbf{BigRNN}}
    & \multicolumn{2}{c}{\textbf{Transformer}}
    \\
    \textbf{\#(Param)} 
    & \multicolumn{2}{c|}{13M}
    & \multicolumn{2}{c|}{37M}
    & \multicolumn{2}{c}{80M}
    \\
    \hline
    &  \textbf{F@5}& \textbf{F@10} & \textbf{F@5}& \textbf{F@10} & \textbf{F@5}& \textbf{F@10}
    \\  
    \hline
    \random 
    & 0.358 & 0.304 & 0.356 & 0.305 & 0.359 & 0.289
    \\
	\length 
    & 0.351 & 0.319 & 0.349 & 0.321 & \textbf{0.361} & 0.318
    \\
	\nosort 
    & \textbf{0.364} & 0.325 & \textbf{0.361} & 0.329 & 0.358 & 0.329
    \\
	\alphab 
    & 0.354 & 0.341 & 0.358 & 0.341 & 0.353 & 0.336
    \\ 
	\appearpre 
    & 0.337 & 0.345 & 0.339 & 0.341 & 0.352 & 0.343
    \\ 
	\appearap 
    & 0.339 & \textbf{0.347} & 0.344 & \textbf{0.346} & 0.357 & \textbf{0.345}
    \\ 
    \hline\hline
  \end{tabular}
  \caption{$F_1$ scores on present keyphrase generation of \onetoseq~models with different model complexities (BeamWidth=50).}
  \label{tab:model_complexity}
  \vspace{-1em}
\end{table}

\subsection{Effects on Absent Keyphrase Generation}
\label{subsec:absent}
Absent keyphrase prediction examines models' abilities to generate synonymous expressions based on the semantic understanding of the text. We report the absent keyphrase results in Table \ref{tab:absent}. Although \onetoseq~models exhibit superior performance in predicting present phrases, they work poorly for absent ones, primarily due to the very limited number of unique predictions they are able to generate. Recall@50 becomes very low as they can hardly produce more than 10 absent phrases. Even with a larger beam width, both RNN models yield very low recall. In contrast, Transformer demonstrates good abstractiveness and beats the baseline on Recall@10. 

\begin{table}[!htbp]
  \centering
  \fontsize{9}{10}\selectfont
  \scriptsize
  \renewcommand{\arraystretch}{1.2}
  \begin{tabular}{c|cc|cc|cc}
    \hline
    \hline
    \textbf{Model} 
    & \multicolumn{2}{c|}{\textbf{BaseRNN}}
    & \multicolumn{2}{c|}{\textbf{BigRNN}}
    & \multicolumn{2}{c}{\textbf{Transformer}}
    \\
    \hline
    & \textbf{R@10}& \textbf{R@50} & \textbf{R@10}& \textbf{R@50} & \textbf{R@10}& \textbf{R@50}
    \\  
    \hline
    \onetoone 
    & \textbf{0.027}	& \textbf{0.068}
    \\
    \hline
    \random 
    & 0.012 & 0.012 & 0.017 & 0.017 & 0.044 & 0.044
    \\
	\length 
    & 0.012 & 0.012 & 0.022 & 0.022 & 0.045 & 0.045
    \\
	\nosort 
    & 0.013 & 0.013 & 0.016 & 0.016 & 0.052 & 0.052
    \\
	\alphab 
    & 0.018 & 0.018 & \textbf{0.026} & \textbf{0.026} & 0.057 & 0.057
    \\ 
	\appearpre 
    & 0.011 & 0.011 & 0.023 & 0.023 & 0.055 & 0.055
    \\ 
	\appearap 
    & 0.015 & 0.015 & \textbf{0.026} & \textbf{0.026} & \textbf{0.070} & \textbf{0.071}
    \\ 
    \hline\hline
  \end{tabular}
  \caption{Recall@10 and @50 on absent keyphrase generation of different models (BeamWidth=50).}
  \label{tab:absent}
  \vspace{-2em}
\end{table}

\section{Conclusion}
\label{section:conclusion}
We present an empirical study on how different orderings affect the performance of \onetoseq~models for keyphrase generation. We conclude our discussion with the following take-aways:
\begin{itemize}
    \setlength\itemsep{-0.1em}
    \item The ordering of concatenating target phrases matters. Consistent with~\citet{vinyals2016_order_matters}, target ordering plays a key role in successfully training models, perhaps due to the potential difference in trainability of each ordering. \appearap~demonstrates the overall best performance among the six ordering we experimented with.
    \item Larger beam width can yield more stable performance boost and reduce the gap among orderings. Model complexity has no significant effect on present keyphrase prediction.
    \item Training with concatenated phrases seems to strengthen a model's extraction capacities more than abstraction capacities. On the other hand, abstraction capacity can be enhanced by increasing model complexity.
    How to balance the two would be a good direction for future study.
    \item By utilizing over-generation (with a beam size of 10), only less than 20\% of phrases generated by \onetoseq~models are unique. Decoding for keyphrase generation remains an open challenge that may deserve more research attention.

\end{itemize}

\bibliography{emnlp-ijcnlp-2019}
\bibliographystyle{acl_natbib}

\appendix

\section{Model Architecture}
\label{appd:structure}

We adopt a sequence-to-sequence based framework with pointer-generator and coverage mechanism proposed by \citet{see17gettothepoint}.
The basic sequence-to-sequence model uses a bi-directional GRU\footnote{For the \textbf{Transformer} experiments described in \S\ref{subsec:complexity}, GRUs are replaced by transformer modules.} \citep{cho14gru} encoder, it takes word embedding vectors of the source text as input and produces a sequence of encoder hidden states, $h_{e}$. On the decoding side of the model, at step $t$, an uni-directional GRU decoder takes the word embedding of the $t-1_{th}$ token as input and generates decoder hidden states $h^{t}_{d}$. Attention mechanism \citep{bahdanau2014attntion} is applied:
\begin{equation}
\begin{aligned}
\mathrm{energy}^{t}&=V^T\tanh(W_eh_e +W_dh^{t}_d + b), \\
\alpha^t&=\smx(\mathrm{energy}^{t}),
\end{aligned}
\end{equation}

where $V$, $W_e$, $W_d$ and $b$ are learnable parameters. $a^t$ represents the probability distribution over correlation between the $t_{th}$ token with each of the source words.
Weighted sum of encoder hidden states,  $h^*_{e}=\sum_i\alpha^{t,i}\cdot h_e^i$, is further used to generate the output $p_{vocab}$ (a probability distribution over an given vocabulary) at step $t$.

The pointer-generator learns a linear interpolation between the two probability distributions (i.e., distribution over source words, $\alpha^t$, and distribution over vocabulary, $p_{vocab}$), so that the combined output considers both distributions. One advantage of pointer-generator is, it can point to out-of-vocabulary tokens when they appear in source text.

As for coverage mechanism, the model maintains a coverage vector $c^t$, which is the sum of attention distributions over all previous decoder time steps. $c^t$ serves as a episodic memory that helps to prevent model from generating repetition.

For more details about the model we use, we refer readers to read \citet{see17gettothepoint}'s clearly described model section.

\section{Generating Variable-number of Phrases}
\label{appd:variable_number_phrase}

One critical advantage of the \onetoseq setting is that the model is capable of generating variable-number of phrases. Specifically, during the decoding phase, we take as the output the multiple phrases from one completely decoded sequence, which usually is the top-ranked sequence (i.e. top beam) from beam search. We refer to this decoding strategy for generating variable-number of phrases as self-terminating generation. 


We report average $F_1$ scores of variable-number keyphrase generation (self-terminating) in Table \ref{tab:topbeam}. We can see that without over-generation, \onetoseq~models are able to achieve a decent performance. A larger beam width results in an improvement on top 10 phrases but a drop on top 5, showing a different phenomenon from \S\ref{subsec:beamwidth}.

\begin{table*}[!htbp]
  \centering
  \fontsize{8}{10}\selectfont
  \renewcommand{\arraystretch}{1.2}
  \begin{tabular}{c|cc|cc|cc|cc}
    \hline
    \hline
    \multirow{2}{*}{\textbf{Method}}
    & \multicolumn{2}{c|}{\textbf{BeamWidth=1}}
    & \multicolumn{2}{c|}{\textbf{BeamWidth=10}}
    & \multicolumn{2}{c|}{\textbf{BeamWidth=25}}
    & \multicolumn{2}{c}{\textbf{BeamWidth=50}}
    \\
    \cline{2-9}
     & \textbf{F$_1$@5} 
     & \textbf{F$_1$@10} 
     & \textbf{F$_1$@5} 
     & \textbf{F$_1$@10} 
     & \textbf{F$_1$@5} 
     & \textbf{F$_1$@10}
     & \textbf{F$_1$@5} 
     & \textbf{F$_1$@10}
    \\  
    \hline\hline
    \random 
    & 0.242 & 0.242 & 0.263 & 0.263 & 0.262 & 0.262 & 0.253 & 0.253
    \\ \hline
	\length 
	& 0.269 & 0.269 & 0.284 & 0.284 & 0.276 & 0.276 & 0.273 & 0.273
    \\ \hline
	\nosort 
	& 0.272 & 0.272 & 0.286 & 0.286 & 0.278 & 0.278 & 0.268 & 0.268
    \\ \hline
	\alphab 
	& 0.270 & 0.270 & \textbf{0.305} & \textbf{0.305} & \textbf{0.292} & \textbf{0.292} & \textbf{0.281} & \textbf{0.281}
    \\  \hline
	\appearpre 
	& 0.274 & 0.274 & 0.280 & 0.280 & 0.275 & 0.275 & 0.265 & 0.265
    \\  \hline
	\appearap 
	& \textbf{0.293} & \textbf{0.294} & 0.300 & 0.300 & \textbf{0.292} & \textbf{0.292} & 0.279 & 0.279
    \\ 
    \hline\hline
  \end{tabular}
  \caption{Experimental results on predicting variable-number of phrases (self-terminating). Best score of each column is highlighted in bold font.}
  \label{tab:topbeam}
\end{table*}

\begin{table}[!h]
  \centering
  \fontsize{9}{10}\selectfont
  \scriptsize
  \renewcommand{\arraystretch}{1.2}
  \begin{tabular}{c|cc|cc|cc}
    \hline
    \hline
    \textbf{Model} 
    & \multicolumn{2}{c|}{\textbf{BaseRNN}}
    & \multicolumn{2}{c|}{\textbf{BigRNN}}
    & \multicolumn{2}{c}{\textbf{Transformer}}
    \\
    \hline
    & \textbf{F@5}& \textbf{F@10} & \textbf{F@5}& \textbf{F@10} & \textbf{F@5}& \textbf{F@10}
    \\  
    \hline
    \onetoone 
    & 0.083	& 0.107	& -- & --& -- & --
    \\	
    \hline
    \random 
    & \textbf{0.147} & 0.154 & \textbf{0.143} & \textbf{0.155} & 0.133 & 0.119
    \\
	\length 
    & 0.140 & 0.156 & 0.138 & 0.139 & \textbf{0.136} & 0.145
    \\
	\nosort 
    & 0.141 & 0.151 & 0.141 & 0.151 & \textbf{0.136} & 0.153
    \\
	\alphab 
    & 0.135 & 0.152 & 0.131 & 0.154 & 0.123 & 0.131
    \\ 
	\appearpre 
    & 0.102 & 0.138 & 0.109 & 0.133 & 0.113 & 0.144
    \\ 
	\appearap 
    & 0.123 & \textbf{0.165} & 0.110 & \textbf{0.155} & 0.133 & \textbf{0.152}
    \\ 
    \hline\hline
  \end{tabular}
  \caption{$F_1$-score on \duc news dataset of different models (BeamWidth=50).}
  \label{tab:duc}
  \vspace{-2em}
\end{table}

\section{Transferring to Another Domain}
Different from the four scientific writing datasets mentioned in \S\ref{section:setting}, \duc~\cite{wan2008single} is a dataset where data is collected from domain of news.
Using \duc, we investigate when transferring to another domain, how different ordering affects an \onetoseq model's test performance.
Table~\ref{tab:duc} presents the results on \duc dataset. 
Note the \duc dataset contains only present phrases, thus most \onetoseq~models outperform the baseline \onetoone~model.

\end{document}